\documentclass[letterpaper, 10 pt, conference]{ieeeconf}  

\IEEEoverridecommandlockouts                              

\usepackage{cite}
\usepackage{mathtools}
\usepackage{amssymb}

\usepackage{multirow}
\usepackage{makecell}
\usepackage{subfigure}
\usepackage{siunitx}

\usepackage{newclude}
\usepackage{booktabs}
\usepackage{color}

\usepackage{soul}

\title{\LARGE \bf
Constrained Reinforcement Learning and Formal Verification\\ for Safe Colonoscopy Navigation}

\author{Davide Corsi$^{1,*}$, Luca Marzari$^{1,*}$, Ameya Pore$^{1,2,*,\dagger}$,\\ Alessandro Farinelli$^{1}$, Alicia Casals$^{2}$, Paolo Fiorini$^{1}$ and Diego Dall'Alba$^{1}$
\thanks{This project has received funding from the European Union’s Horizon 2020 research and innovation programme under the Marie Skłodowska-Curie (grant agreement No. 813782 "ATLAS")}
\thanks{* These authors contributed equally to the paper. Ordered alphabetically.}
\thanks{$^{1}$ Department of Computer Science, University of Verona, Italy}
\thanks{$^{2}$ Research Centre for Biomedical Engineering (CREB), Technical University of Catalonia (UPC), Spain}
\thanks{$\dagger$ Corresponding author: Ameya Pore (email: ameya.pore@univr.it)}}

\begin{document}

\maketitle
\thispagestyle{empty}
\pagestyle{empty}

\begin{abstract}
The field of robotic Flexible Endoscopes (FEs) has progressed significantly, offering a promising solution to reduce patient discomfort. However, the limited autonomy of most robotic FEs results in non-intuitive and challenging manoeuvres, constraining their application in clinical settings. While previous studies have employed lumen tracking for autonomous navigation, they fail to adapt to the presence of obstructions and sharp turns when the endoscope faces the colon wall. In this work, we propose a Deep Reinforcement Learning (DRL)-based navigation strategy that eliminates the need for lumen tracking. However, the use of DRL methods poses safety risks as they do not account for potential hazards associated with the actions taken. 
To ensure safety, we exploit a Constrained Reinforcement Learning (CRL) method to restrict the policy in a predefined safety regime.
Moreover, we present a model selection strategy that utilises Formal Verification (FV) to choose a policy that is entirely safe before deployment. We validate our approach in a virtual colonoscopy environment and report that out of the 300 trained policies, we could identify three policies that are entirely safe. Our work demonstrates that CRL, combined with model selection through FV, can improve the robustness and safety of robotic behaviour in surgical applications.

\end{abstract}

\section{INTRODUCTION}

Early detection of ColoRectal Cancer (CRC) is a key element for achieving optimal disease treatment and improving the survival rate.
Colonoscopy is a widely adopted diagnostic and therapeutic procedure for CRC, where a Flexible Endoscope (FE) is manually operated by expert interventionist \cite{manfredi2021endorobots}.
However, there are two major drawbacks to this procedure. Firstly, patients without sedation experience significant discomfort and pain due to tissue stretching associated with FE manipulation.
The most frequent cause of pain is looping, where the FE advances into the colon without a corresponding progression of the tip. Looping also increases the risk of colon perforation and massive bleeding \cite{manfredi2021endorobots}. Secondly, the non-intuitive and challenging manipulation of FE necessitates highly skilled clinicians. A shortage of endoscopists relative to the clinical demand leads to increased physical strain and musculoskeletal injuries \cite{martin2020enabling}.

\begin{figure}[t]
	\centering
	\includegraphics[width=\linewidth]{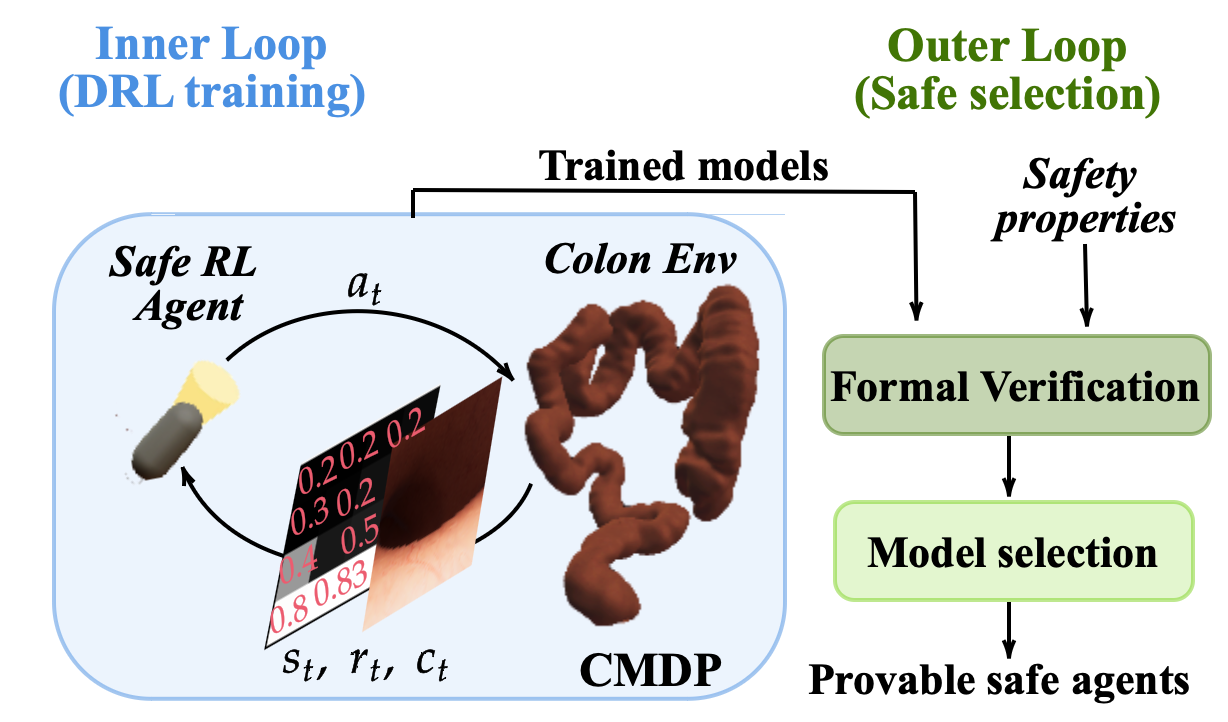}
	\caption{Safe-Reinforcement learning framework proposed in this work. Agents are trained in a CMDP setting with soft constraints. The trained policies are examined with the FV tool which identifies the safety violations. Policies without safety violations are selected for final deployment to ensure a completely safe behavior.
    }
	\label{fig:title_image}
 \vspace{-4mm}
\end{figure}

Robotic Endoscopes (RE) provide a less painful and more ergonomic approach to colonoscopy. 
However, manual control of RE is prone to human error and requires extensive operator training. These limitations have motivated the development of autonomous navigation systems since navigation is one of the principal phases of colonoscopy \cite{pore2023autonomous}.


Autonomous navigation systems based on visual information 
use different processing techniques, which
rely on the assumption that the region of maximum depth within an image represents a valuable target for immediate heading adjustment. 
This region corresponds with the darkest area in the endoscopic image. Thus, 
different segmentation methods for the estimation of darkest region have been proposed based on contour estimation \cite{prendergast2018autonomous, prendergast2020real}, optical flow \cite{reilink2010image}, image intensity \cite{martin2020enabling} and convolutional neural network (CNN) \cite{lazo2022autonomous}.

Regardless of the estimation method used, once the deeper or darker region is detected, a rule-based controller, commonly based on Proportional-Integral-Derivative approach \cite{lazo2022autonomous, martin2020enabling, prendergast2018autonomous} or finite state machines \cite{prendergast2020real}, is used to minimize the error with respect to the center of the endoscopic image, called lumen distance.
These controllers are not robust to rapid changes in the estimates provided, often due to errors in the segmentation method or dynamic deformations of the anatomy.
As an alternative, Deep Reinforcement Learning (DRL) has been proposed for generating adaptive control signals, providing an end-to-end mapping between the endoscopic images and the endoscope's control signal \cite{pore2022colonoscopy}. The DRL method requires a reward function that minimizes the lumen distance, which itself requires lumen detection.
In this work, we focus on developing an end-to-end DRL approach that is independent from a separate perception system, similar to \cite{li2023rl}. 

Nevertheless, implementing DRL in real-world robotic systems raises concerns over safety, which is of utmost priority in surgical settings. DRL methods are susceptible to unanticipated behaviors in situations not encountered during training, leading to potentially harmful consequences \cite{AcHeTa17}.

Constrained Reinforcement Learning (CRL) provides a way to tackle safety by restricting the agents from taking potentially unsafe actions through the incorporation of an additional cost function that should be minimized. While the reward function incentivizes specific behavior, the cost function is designed to penalize unwanted actions. However, in practice, achieving a perfect zero-cost result through numerical optimization in DRL is often impractical. Thus, a threshold is set as a maximum acceptable value for the cumulative cost. 
Examples of algorithms to face this challenging problem include 
CPO \cite{AcHeTa17}, based on the concept of safe policy improvement or SOS \cite{MaCoFa21b}, which incorporates a genetic step in the training loop. 
In this paper, we focus on the Lagrangian Proximal Policy Optimization (L-PPO) algorithm, which leverages the Lagrangian dual relaxation of a constrained optimization problem \cite{RaAcAm19}. 
L-PPO inherits all the strengths of the PPO algorithm (e.g., trust-region policy improvement and first-order optimization) while offering a simple and efficient method for updating constraints.

In CRL methods, safety specifications estimate the expected risk over the entire trajectory and do not guarantee safety at a particular state \cite{LiuSurvey, AcHeTa17}.
Hence, ensuring that the DRL agent never causes safety violations is crucial.
Formal Verification (FV) has been identified in the literature as a mathematical framework to establish safety in DRL systems \cite{ProVe, TACAS} and has been recently applied to robot-assisted surgical tasks \cite{pore2021safe}.
However, previous FV approaches applied to robot-assisted surgical setups only identify states that can potentially lead to safety violations without utilizing the findings for other scalable objectives, such as model selection.
In this study, we develop a model selection strategy that chooses policies without any safety violation, by formally verifying each of the policies over a set of safety properties.

Therefore, we propose a novel framework that integrates the following features:
(1) An end-to-end DRL method for colon navigation that eliminates the need for a separate lumen detection system.
(2) A CRL approach that constrains the policy in a pre-defined safe state-space to minimize potentially dangerous actions. 
(3) A model selection strategy that selects policies satisfying all safety constraints, with each policy formally verified to check for its safety violations.

We evaluate the proposed framework in a virtually simulated colonoscopy setup that accurately emulates the dynamics of colon tissue. 
The colon navigation performance and the safety of the proposed CRL approach is evaluated against the standard DRL approach.
This work demonstrates that the combination of CRL and FV can improve safety in autonomous colonoscopy navigation.

\section{Problem Statement}
In this section, we give an overview of the colonoscopy environment used and briefly introduce the safety objectives for autonomous navigation.

\subsection{Colonoscopy Environment}

The 3D models of the colon are derived from publicly available CT colonography datasets and are refined to generate volumetric and superficial meshes with realistic textures \cite{incetan2021vr,pore2022colonoscopy}. In order to simulate the deformable nature of the colon, a biomechanical model based on the Simulation Open Framework Architecture (SOFA) is integrated in Unity3D to obtain high quality realistic anatomical environment.

To simulate the navigation of a RE tip actuation, we consider a rigid capsule with a camera, weighing 20g and with a length and diameter of 36mm and 14mm, respectively, as the endoscope tip. This modeling approach is consistent with previous works \cite{incetan2021vr,pore2022colonoscopy}.

\subsection{Overview of the safety framework}

\begin{figure}[thpb]
	\centering
	\includegraphics[width=\linewidth]{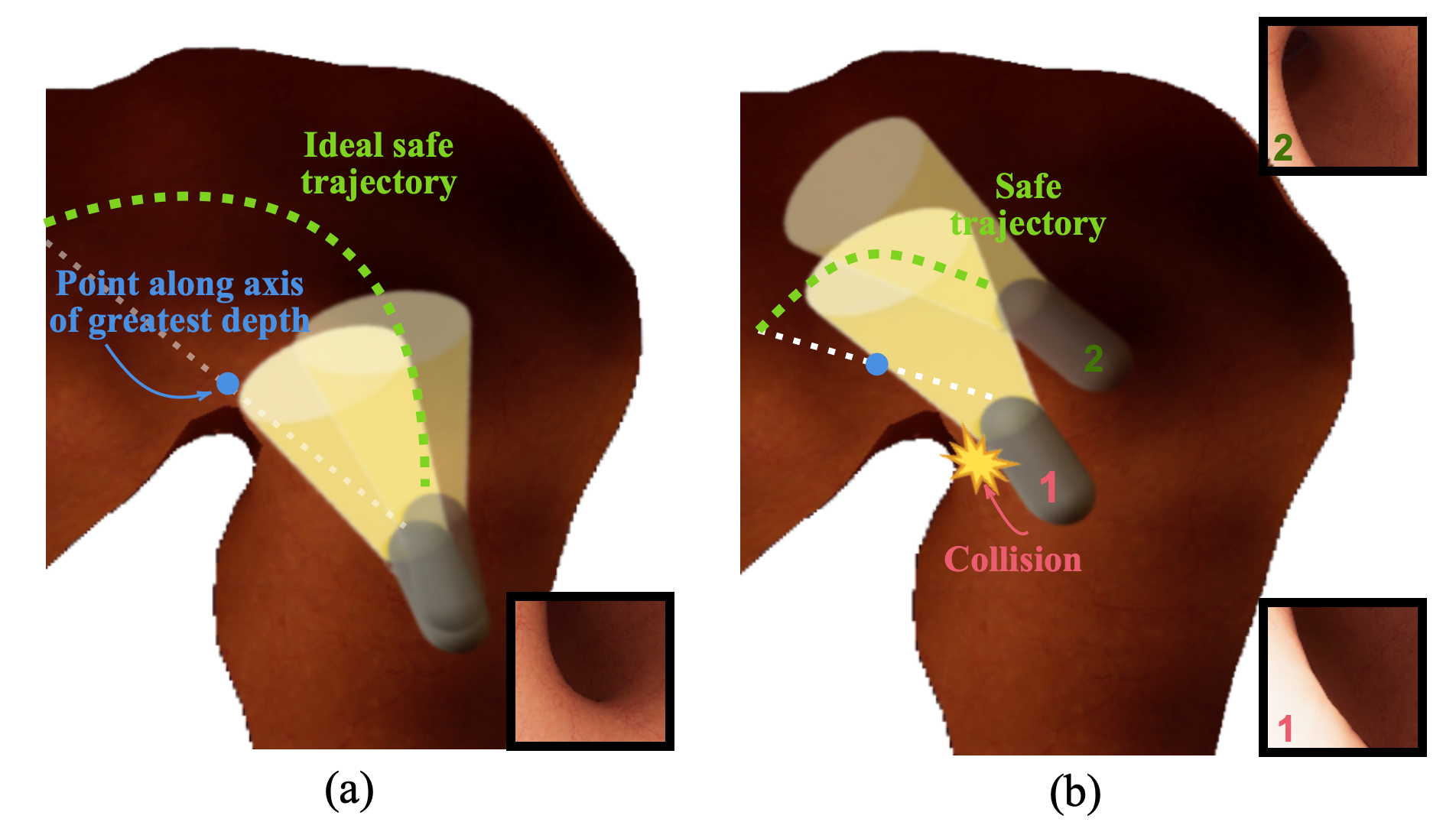}
     \vspace*{-8mm}
	\caption{Capsule endoscope positioned inside the lumen, facing an upcoming turn, with both the point of greatest depth (darkest point) and the lumen center visible.  If the capsule endoscope is guided towards the deepest point, it will inevitably cause biased motion towards the inner wall of the turn, as shown in (a). This biased motion could potentially lead to occlusion of the camera and collision with the wall, as shown in (b). The right-hand side square boxes display the endoscopic view, with (a) showing the two endoscope tips at a similar position, providing the same view, and (b) illustrating that endoscope 1 approaches the wall more closely than endoscope 2 as it follows the line of greatest depth.
  }
	\label{fig:image_segmentation}
 \vspace{-3mm}
\end{figure}

The conventional approach of using the region of greatest depth as the immediate heading adjustment goal suffers from an intrinsic weakness, where lighting conditions, focal length, and surrounding tissue geometry can considerably impact the actual distance to the deepest point, making it an unreliable target for precise navigation \cite{lazo2022autonomous, martin2020enabling, prendergast2018autonomous}. 
Relying solely on the perfect alignment of the camera towards the deepest point ignores the 3D structure of the surrounding anatomy and will inherently limit the ability of the endoscope to navigate through tight bends (as depicted in Fig.~\ref{fig:image_segmentation}), where a large proportion of images will not be well-centered within the lumen (shown in Fig.~\ref{fig:observation_space}a). This results in close-up views of the lumen wall, which can be highly illuminated due to reflection. Thus, defining safety objectives to prevent the endoscope from moving in the orthogonal direction of the colon wall, which could potentially lead to perforation, can result in a safer trajectory.

Henceforth, we establish two safety indices for our study:
(1) Soft constraints which provide guidance for avoiding colon wall collisions based on safety probability analysis and optimization. The incorporation of soft constraints during the training of standard DRL methods facilitates the agent to learn actions that conform to safe configurations, leading to CRL.
(2) Hard constraints which impose strict restrictions on the system to prevent it from entering specified unsafe regions, such as perforation. We observe that movement of the scope towards the illuminated region of the image can lead to a trajectory orthogonal to the wall. Consequently, we propose a set of four hard constraints, referred to as safety properties ($\Theta$), 
based on user-defined brightness thresholds in different image regions.
If the values surpass the threshold, the robot must restrict its actions in that direction.

It is often difficult for CRL methods to enforce hard constraints by setting indirect constraints on the expectation of cumulative cost \cite{LiuSurvey, AcHeTa17}. 
Therefore, a key objective of our work is to leverage FV techniques to analyze the policy and assess its adherence to the specified hard constraints.

\section{Constrained Reinforcement learning} \label{sec:constrained_RL}

\subsection{Deep Reinforcement Learning (DRL)}
The problem of colon navigation is formulated as a Markov Decision Process (MDP), which can be represented as a tuple $(\mathcal{S}, \mathcal{A}, \mathcal{R}, \mathcal{T}, \gamma, H)$. Here, $\mathcal{S}$ denotes the state space, $\mathcal{A}$ is the action space, $\mathcal{T}$ is the transition probability distribution, $\mathcal{R}$ is the reward space, $\gamma \in [0, 1]$ is the discount factor, and $H$ is the time horizon per episode. At each time step $t$, the environment produces a state observation $s_{t} \in \mathcal{S}$, and the agent generates an action $a_t \in \mathcal{A}$ according to a policy $a_t \sim \pi(s_t)$. The agent then applies this action to the environment, receiving a reward $r_t \in \mathcal{R}$. This process leads to the agent transitioning to a new state $s_{t+1}$ sampled from the transition function $p(s_{t+1} | s_{t}, a_{t})$, where $p \in \mathcal{P}$, or the episode terminates at state $s_{H}$. The objective of a DRL algorithm is to find a policy that maximizes the expected cumulative reward over a trajectory. This can be expressed as:
\vspace{-1mm}
\begin{equation}
	\label{eq:maximization_drl}
	\max_{\pi_\theta \in \Pi} J_{r}({\pi_\theta}) := \mathbb{E}_{\tau \sim \pi_\theta}[R(\tau)]
 \vspace{-1mm} 
\end{equation}
Here, $\pi_\theta$ represents a policy parameterized by $\theta$, $\tau$ denotes a trajectory, and $R(\tau)$ denotes the cumulative reward obtained along that trajectory.

In literature, different approaches exist to solve this problem (e.g., \cite{HaZhTu18}). However, in this paper, we emphasise on Proximal Policy Optimization (PPO), which is widely acknowledged as the state-of-the-art algorithm and one of the most efficient methods for control problems \cite{schulman2017proximal}. 

\subsubsection{Observation Space}
\begin{figure}[t]
    \centering
    \includegraphics[width=\linewidth]{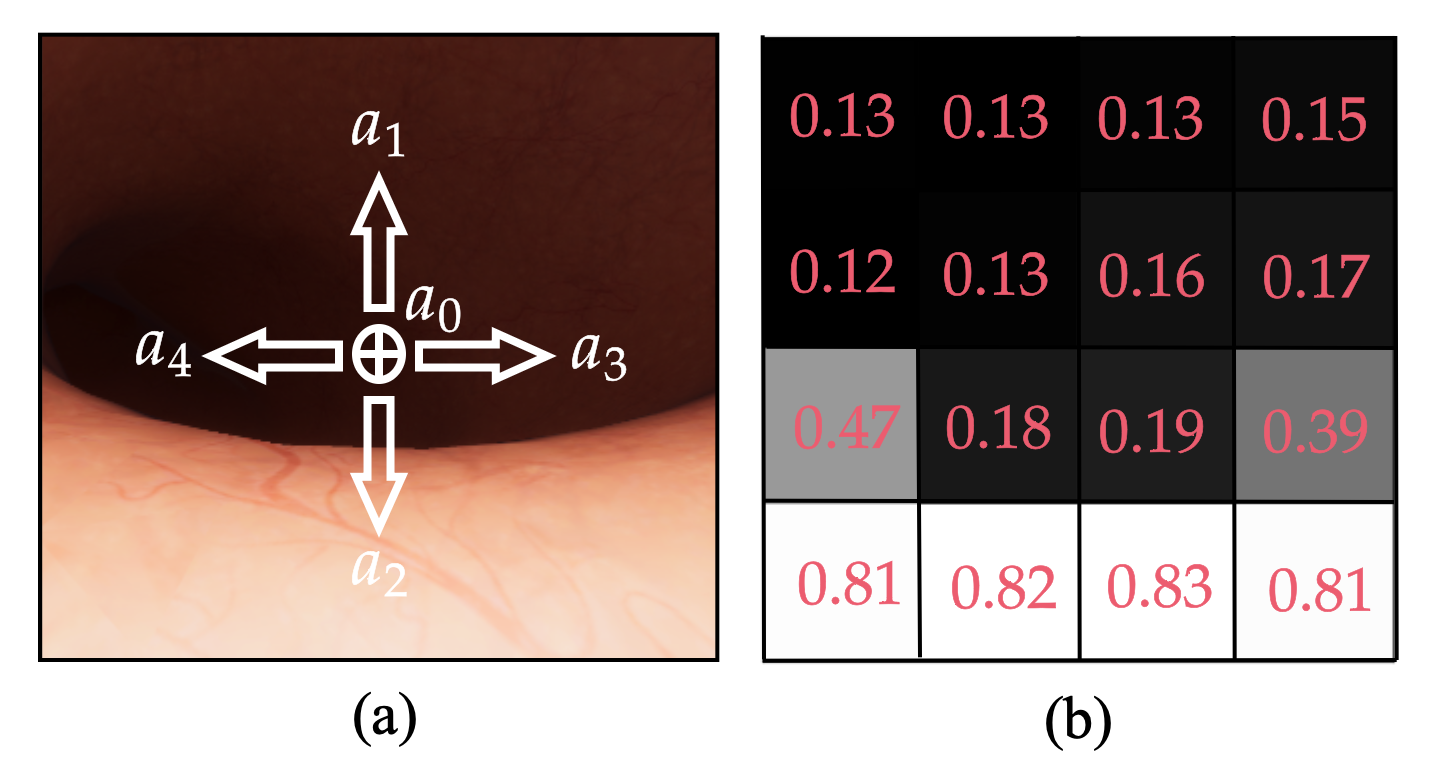}
    \vspace*{-5mm}
    \caption{(a) Endoscopic view with the allowed actions. (b) Discrete representation of the input space used for the agent.}
    \label{fig:observation_space}
    \vspace{-5mm}
\end{figure}

: 
The observation space is characterized by a low-dimensional discretization of the endoscopic image, represented by a 4x4 matrix (as depicted in Fig.~\ref{fig:observation_space}b). 
The image is first discretized by dividing each dimension into four regions, and each square region is assigned a value that represents the normalized average of the underlying pixels. 
We demonstrate that achieving state-of-the-art autonomous navigation performance is feasible even with a discretized low-dimensional input space, without relying on CNN or other complex architectures.
This discretization step further simplifies the FV process which can be computationally expensive and face scalability challenges when dealing with high-dimensional inputs like images \cite{ProVe, CountingProVe}.
The resulting 2-D down-scaled image is then flattened into a list of 16 values, which form the input to the DRL algorithm.

\subsubsection{Action space}: 
The action space is comprised of five discrete actions, each of which corresponds to a movement in one of the four cardinal directions, namely $a_1:$\texttt{up}, $a_2:$\texttt{down}, $a_3:$\texttt{right}, and $a_4:$\texttt{left}; plus an additional action, $a_0:$\texttt{center}, to set the angular velocity to zero. The agent moves at a constant linear velocity of \SI{3}{\milli\meter/\second}. The angular velocity, which determines the rotation of the endoscope tip, is dependent on the specific action selected and corresponds to a fixed angle of \SI{0.017}{\radian/\second} in the two degrees of freedom. To facilitate the input-output mapping, the neural network controller has been designed with 5 output neurons (one for each action) and 16 input nodes, which is consistent with the discretized image representation discussed in the previous subsection.

\subsubsection{Reward function}: 
We design a reward function that incentivizes the agent to reduce the distance from the end of the colon while minimizing the interactions with the colon wall. In light of this, we have formulated a reward function that provides a high positive reward to the agent upon successful completion of the task, a small penalty upon touching the colon wall, and an additional penalty that scales with the distance between the agent and the end of the colon. The mathematical expression of the reward function is as follows:
\vspace{-1mm}
\begin{equation}
\begin{small}
    R_t = \begin{cases}
    10 \hfill \text{reaches the end} \\
    -\beta \hfill \text{touches the wall} \\
    (-dist_{t}) \cdot \eta \hspace{30px} \text{otherwise}\\
    \end{cases}
\label{eq:methods:reward_function}
\end{small}
\end{equation}

where $dist_t$ is the centerline distance from the end of the colon at time $t$. The centerline distance for each colon model is estimated prior to training using checkpoints. 
$\eta$ is a normalization factor, and $\beta$ is a fixed penalty for each collision. The values of $\eta$ and $\beta$ are empirically set to 0.001 and 0.01, respectively, in our experiments.

\subsection{Constrained DRL and Lagrangian-PPO}
In the previous sections, we have discussed the concept of optimal policy for an MDP. However, in safety-critical scenarios, it is necessary for an agent to guarantee additional behaviors of paramount importance, even more than achieving the primary task \cite{RaAcAm19}. For instance, in colonoscopy, preventing lumen wall perforation takes precedence over reaching the destination, despite the latter being the primary objective \cite{prendergast2020real}.

This issue is typically addressed by modeling the problem using a Constrained Markov Decision Process (CMDP), which is an extension of a standard MDP that includes an additional signal, namely the \textit{cost function}, defined as $C: \mathcal{S} \times \mathcal{A} \rightarrow \mathbb{R}$, and a threshold value $d \in \mathbb{R}$ that the expected value of the cost must remain below. For the sake of simplicity, we consider the case of only one cost function and its corresponding threshold, but the framework can be easily extended to handle multiple constraints. We formally define the set of feasible policies for a CMDP as follows:
\vspace{-2pt}
\begin{equation}
	\Pi_\mathcal{C} := \{\pi_\theta \in \Pi : \ \forall k, \ J_{C}(\pi_\theta) \leq d\}
	\label{eq:background:valid_cmdp}
 \vspace{-1mm}
\end{equation} 

\noindent where $J_{C}(\pi_\theta)$ is the expected cost function over the trajectory and $d_k$ is the corresponding threshold. 

A constrained DRL algorithm should find a policy $\theta \in \Pi_\mathcal{C}$ that maximizes the reward. A natural way to encode this problem is through a constrained optimization problem in the form of: 
\vspace{-2pt}
\begin{equation}
	\begin{aligned}
		\max_{\pi_\theta} \quad & J_{r}({\pi_\theta}),
		\quad \textrm{s.t.} \quad &  J_{C}(\pi_\theta) \leq d\ \\
	\end{aligned}
 \vspace{-1mm}
\end{equation}
One viable method to incorporate the constraints in an optimization problem involves the utilization of \textit{Lagrange multipliers}. In the context of DRL, a possible technique is to transform the constrained problem into its dual unconstrained counterpart. The objective function for optimization can be expressed as follows: 
\vspace{-2pt}
\begin{equation}
	J(\theta) = \min_{\pi_\theta} \max_{\lambda \geq 0} \mathcal{L}(\pi_\theta, \lambda)
 \vspace{-1mm}
\end{equation}

\noindent where $\mathcal{L}(\pi_\theta, \lambda) = J_r(\pi_\theta) - \lambda(J_{C}(\pi_\theta) - d)$. Among the various DRL algorithm that can be used to maximize this function, a typical choice is to exploit PPO, which has shown promising results when applied together with the Lagrangian dual optimization \cite{RaAcAm19}.
In our setup, we consider the same reward function of PPO, with the addition of a cost function that assumes a value of $1$ only when the capsule interacts with the wall; otherwise is always $0$. 
A cost threshold values of $500$ was selected, balancing the safety of the capsule without compromising its reward performance (discussed in Sec.\ref{sec:exp}).
\section{FORMAL VERIFICATION} \label{sec: fv}

\begin{figure*}[]
    \centering
    \includegraphics[width=0.9\linewidth]{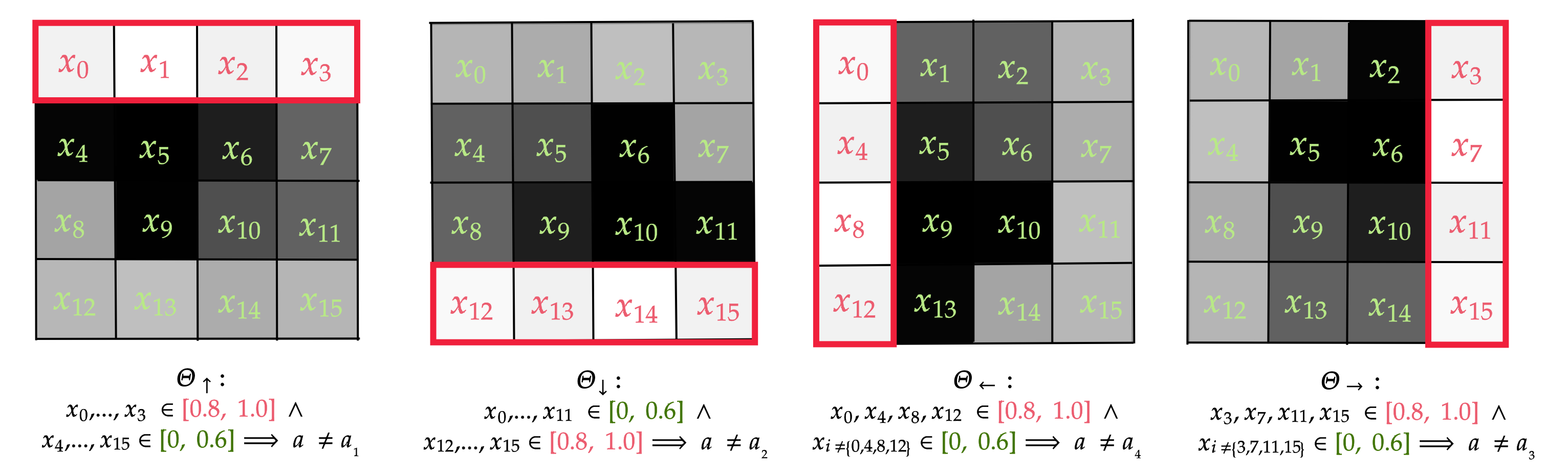}
    \caption{Illustration of four safety properties designed, namely $\Theta_{\downarrow}, \Theta_{\uparrow}, \Theta_{\rightarrow},$ and $\Theta_{\leftarrow}$ . 
    When the scope is close to the upper, lower, left, or right lumen wall, the respective row squares in the input space have high illumination with values in $[0.8,1]$, hence the agent should not move in that direction.
    }
    \label{fig:safety_properties}
    \vspace{-5mm}
\end{figure*}


The FV of DNNs \cite{LiuSurvey, Reluplex, CountingProVe} 
is mathematically defined by the tuple $\mathcal{R}=\langle\mathcal{F}, \mathcal{P}, \mathcal{Q}\rangle$, where $\mathcal{F}$ is a trained DNN, $\mathcal{P}$ is a precondition on the input, and $\mathcal{Q}$ is a postcondition on the output. The precondition $\mathcal{P}$ specifies the admissible input configurations that are of interest, while the postcondition $\mathcal{Q}$ represents the desired output results that must be verified. 
Solving the verification problem requires demonstrating the existence of at least one concrete input (vector) $\Vec{x}$ that satisfies the given constraint, as formulated by the following assertion: 
\begin{equation}
\exists\;\Vec{x}\;\vert\;\mathcal{P}(\Vec{x}) \wedge \mathcal{Q}(\mathcal{F}(\Vec{x}))
\end{equation}
The verification algorithm employs a search procedure to determine if an input vector $\Vec{x}$ satisfying the precondition $\mathcal{P}$ and the postcondition $\mathcal{Q}$ exists and returns $\texttt{SAT}$ if it does \cite{Reluplex}.
We employ VeriNet, a state-of-the-art FV tool \cite{Verinet2} to solve the verification problem.

In this work, we define a set of safety properties, denoted as $\Theta_{\downarrow}, \Theta_{\uparrow}, \Theta_{\leftarrow}$ and $\Theta_{\rightarrow}$, to ensure safe operation of the agent during colonoscopy, shown in Fig.~\ref{fig:safety_properties}. The safety properties are expressed using the possible input values $x_0,...,x_{15} \in \Vec{x}$ of $\mathcal{F}$, where $x_0$ and $x_{15}$ represent the values of the upper-left and bottom-right squares, respectively and the five possible actions, denoted as 
$a_0, ..., a_4$ (highlighted in Fig.~\ref{fig:observation_space}a), that the agent can take.

The precondition $\mathcal{P}$ is encoded using a set of hyper-rectangles, represented by intervals, one for each possible input value.
In more detail, we consider two types of intervals to encode $\mathcal{P}$: $[0, 0.6]$ represents a safe image area, free of obstacles, while the interval $[0.8, 1]$ represents a bright image area, i.e., the agent is close to the colon wall (illustrated in Fig.\ref{fig:safety_properties}). 
The postcondition $\mathcal{Q}$ requires the agent to choose any action other than $a_i$, which corresponds to the unsafe action of scope motion in the direction of illumination. Hence, to verify these properties, FV searches for a single input $\Vec{x}$ that satisfies $\mathcal{P}$ and for which $\mathcal{F}$ satisfies the negation of the postcondition, i.e., a configuration in which the agent selects the unsafe action $a_i$. If no such configuration is found, the original property holds.

It is important to emphasize that the safety properties outlined in our study specify the action the agent should not take in an unsafe situation. However, they do not specify which action should be taken instead. This is a crucial concept because we do not want to force the agent to select a specific action, limiting its capability of finding novel and optimal strategies, but instead only avoiding the most harmful actions.

\section{EXPERIMENTAL VALIDATION}  \label{sec:exp}

In this section, we present the results of the empirical evaluation of the proposed framework. The experiments aim to address the following research questions: (Q1) \textit{What is the effect of a constrained approach on the training of the agent and its performance in the task of autonomous colon navigation?} (Q2) \textit{Can reducing the violation of soft constraints lead to the elimination of violations of hard constraints?}

\subsection{Experimental setup}
The evaluation of the proposed framework is based on
four colon models of varying complexity, characterized by their length and the number of acute bending angles (exceeding 90$^{\circ}$), as detailed in \cite{pore2022colonoscopy}. 

The primary objective is to evaluate the differences in training the proposed CRL (L-PPO) and standard DRL (PPO) approaches. The following steps were used for evaluation.
(S1) 5000 policies were trained on the hardest colon model with different random initialization.
(S2) The best 300 policies were selected based on success rate during training, which is the number of times the agent successfully reaches the colon end in 100 consecutive trials while minimizing the number of collisions with the walls.
(S3) These 300 policies were evaluated on other colon models, and the navigation performance based on the average distance traveled by the scope on each colon model was recorded. 
(S4) FV was performed on the 300 policies to obtain a policy that shows no safety violation for final deployment.
All data were obtained using an RTX 2070 and an i7-9700k.

While carrying out S3, in addition to the considered methods (i.e. L-PPO and PPO), we also include the results of the PPO$_{lum}$ method in Table~\ref{tab:centerline_results}. PPO$_{lum}$ represents the PPO baseline trained using the lumen centralization reward function proposed in prior research by \cite{pore2022colonoscopy}. The average distance travelled is a crucial factor in evaluating trajectories since multiple backward motions or reversing the direction of motion may lead to suboptimal trajectories. The measurement of distance travelled utilizes position values of the endoscope tip normalized by the centerline distance of the colon model. 
\vspace{-1mm}
\begin{table}[h!]
\centering
\caption{Average distance traveled results.}
\vspace{-2mm}
\begin{tabular}{|c|c|c|c|c|}
\hline
                             & \textbf{Colon 0} & \textbf{Colon 1} & \textbf{Colon 2} & \textbf{Colon 3} \\ \hline
\textbf{PPO$_{lum}$ \cite{pore2022colonoscopy}} & \textbf{0.84}            & 0.85          & 0.97            & 0.92         \\ \hline
\textbf{PPO}            & 0.86              & 0.92             & 0.99            & 0.91             \\ \hline
\textbf{L-PPO}               & 0.88              & \textbf{0.81}               & \textbf{0.92 }            & \textbf{0.84}             \\ \hline
\end{tabular}
\label{tab:centerline_results}
\vspace{-3mm}
\end{table}


\begin{figure}[h]
\centering
\begin{center}
    \subfigure[Expected returns] {
    \includegraphics[width=0.45\linewidth]{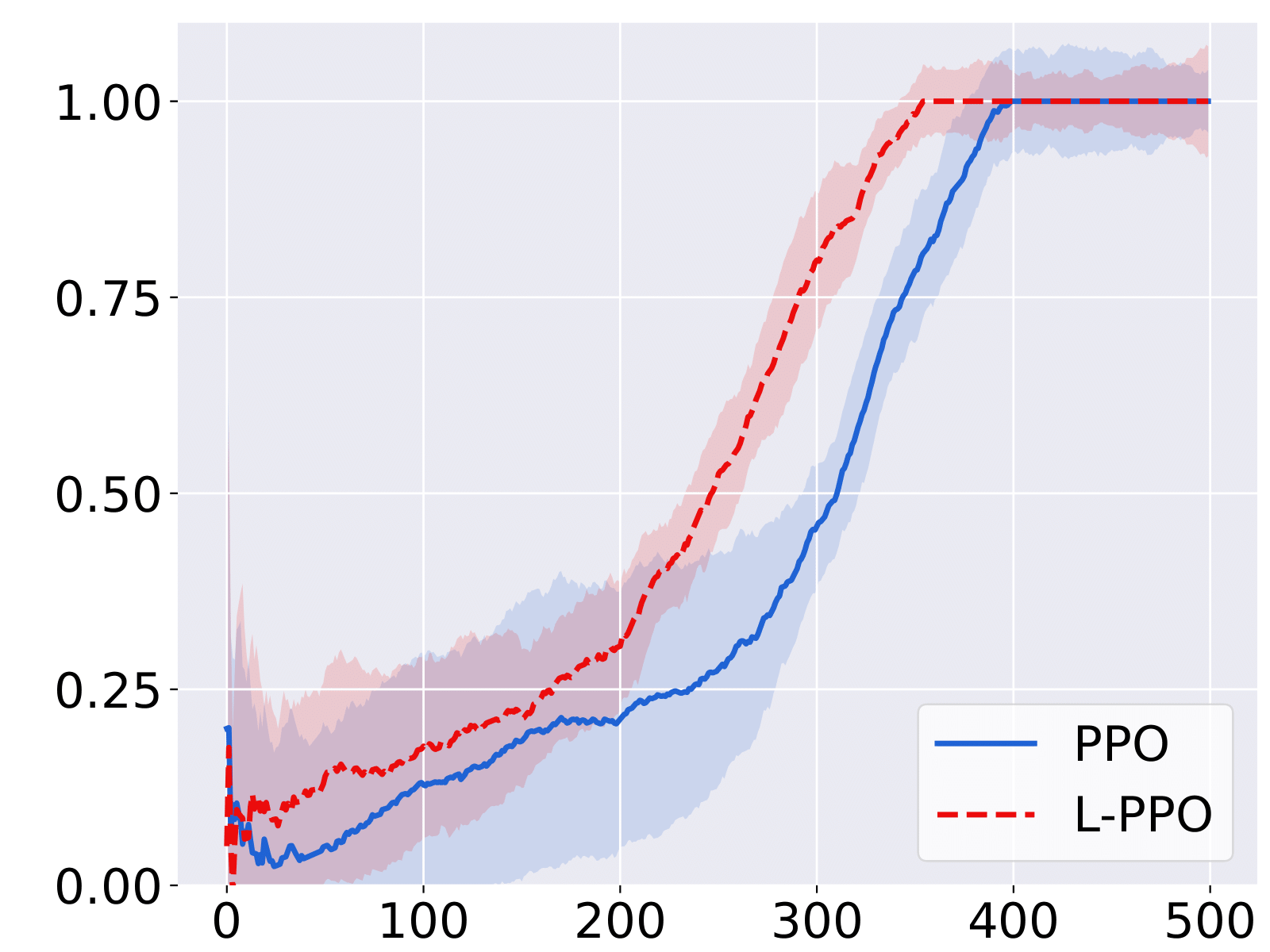}
    }
    \subfigure[Cumulative Cost] {
    \includegraphics[width=0.45\linewidth]{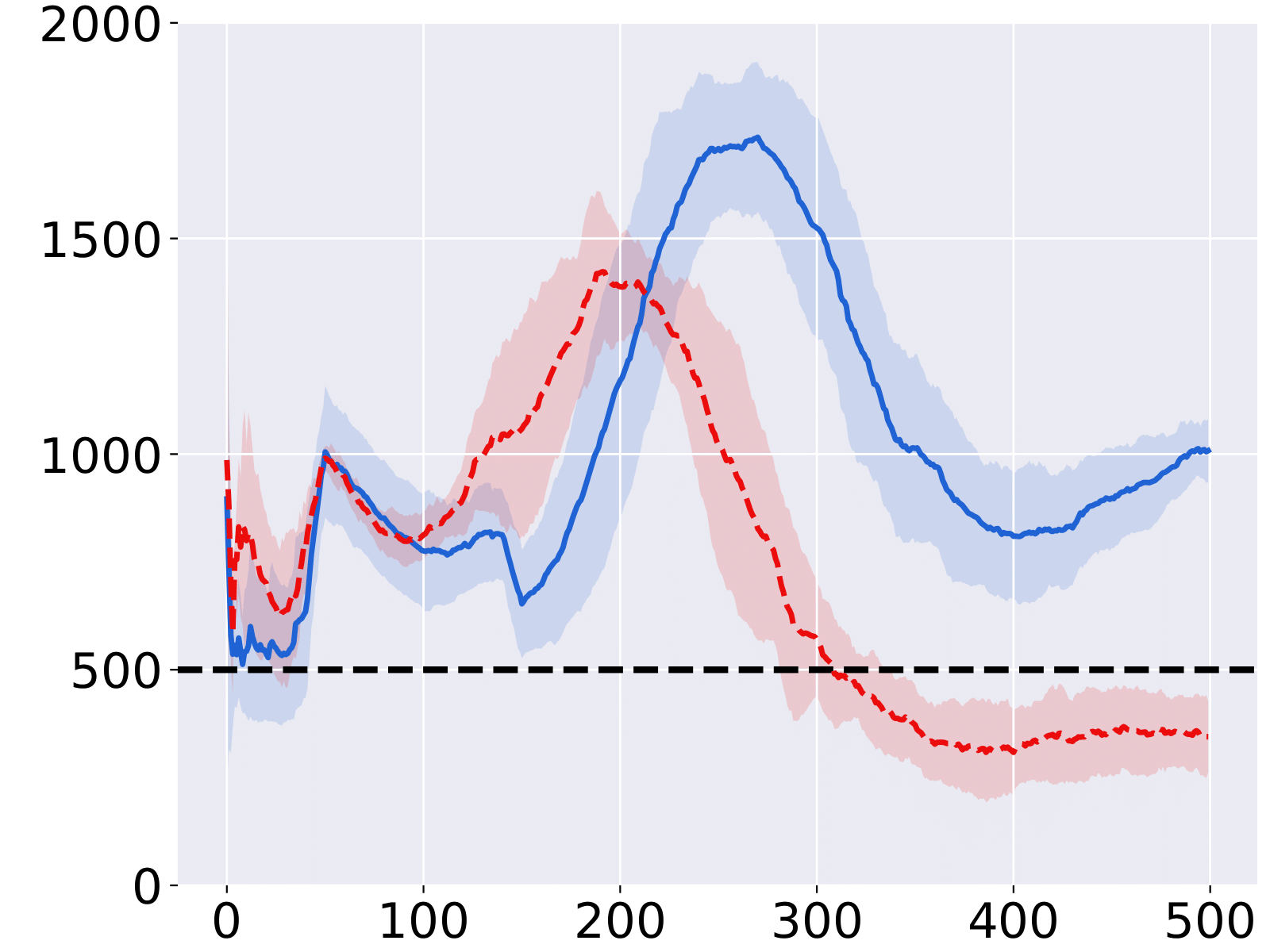}
    }
\end{center}
\vspace{-10pt}
\caption{Average performance vs the number of episodes of PPO and L-PPO over ten seeds (a) Expected returns and (b) Cumulative cost. 
Solid blue and red dashed lines are the empirical mean, while shaded regions represent the standard deviation. Black dashed line is the cost threshold}
\label{fig:training_results}
\vspace{-5mm}
\end{figure}

\subsection{Training results}

The learning curves of PPO and L-PPO are presented in Fig.~\ref{fig:training_results}a, demonstrating a comparable performance between the two algorithms, with both reaching higher reward values at approximately 400 episodes. Our analysis reveals that L-PPO effectively enforces constraints by maintaining a constraint cost below the limit value at 300 episodes, while PPO's constraint cost remains above the limit. Note that a single collision can produce a large number of interactions, depending on the number of timesteps the agent stays in contact with the wall. These results suggest that L-PPO can achieve better constraint satisfaction on average than PPO.
Our examination further demonstrates that both PPO and L-PPO achieve a 100\% success rate in navigating all colon models by reaching the end of the colon. 

We observe an equivalent performance among all three algorithms, indicating that DRL can be trained without a lumen centralization reward, over a global objective of reaching the colon end. 
The results in Table~\ref{tab:centerline_results} indicate that all three algorithms follow a path shorter than the centerline. L-PPO shows the shortest path for Colon 1, 2, and 3. As for Colon 0, which represents a simple scenario, all three algorithms perform well, making it difficult to determine the cause of L-PPO's lower performance on Colon 0.


\subsection{Formal verification results} \label{sec:results}


To address Q2, FV is conducted on the 300 policies trained using each methodology. Table~\ref{tab:formal_verification} provides the violations for PPO and L-PPO across all four safety properties. Specifically, for each safety property, 
we report the $\texttt{SAT}$ values indicating the number of models that violate that particular property.
Notably, we observe that for the first safety property $\Theta_\uparrow$, which pertains to the situation where the upper part of the image is very bright and does not require an upward action from the agent, all 300 PPO policies violate the safety property. The observed violation is not straightforward to interpret, and it may be attributed to the infrequent exposure of the agent to such setups during the training process. It is plausible to suggest that the lack of sufficient training data for these specific scenarios may have hindered the agent's ability to learn the corresponding actions that adhere to the prescribed safety property.

\begin{figure}[h!]
	\centering
	\includegraphics[width=\linewidth]{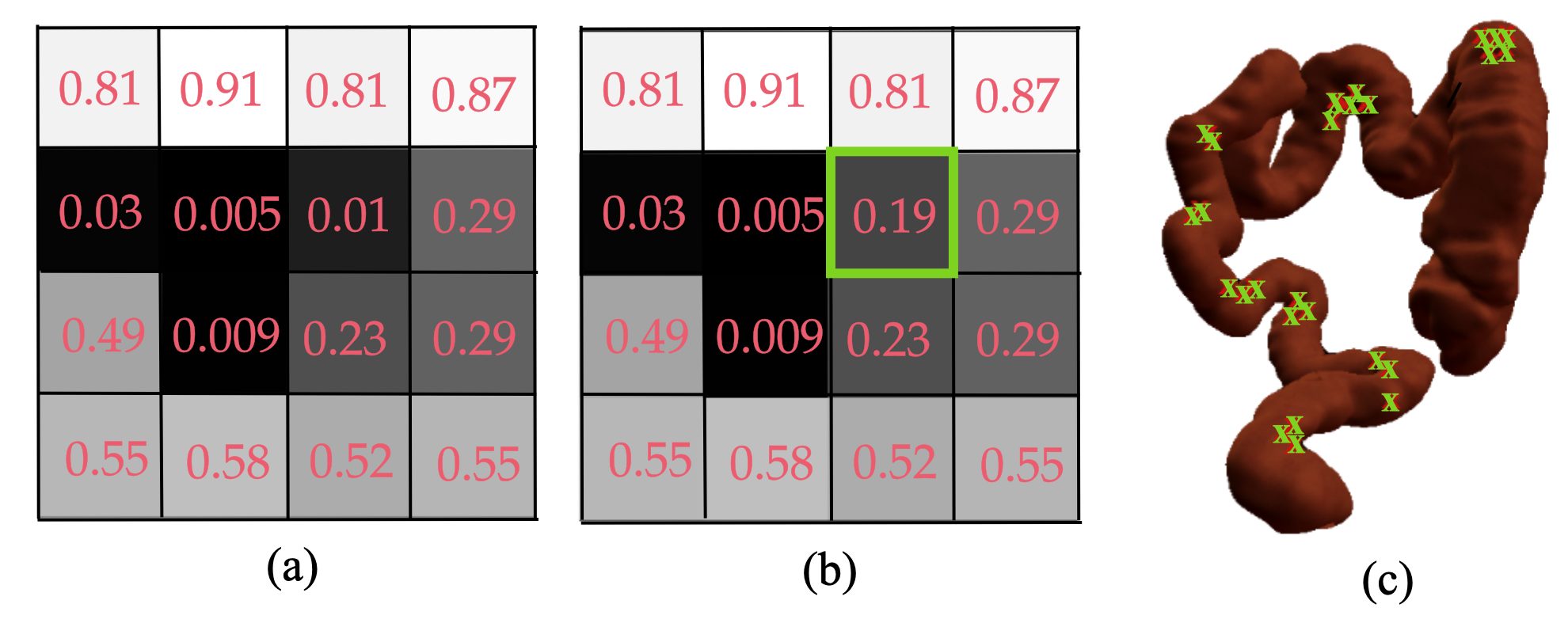}
    \vspace*{-7mm}
	\caption{(a) Adversarial example discovered with FV for the safety property $\Theta_{\uparrow}$. (b) A small perturbation in the square marked green, the agent shows safe behavior. (c) Hard constraint violation positions for one of the PPO policies marked with green crosses.}
	\label{fig:adversarial}
\vspace{-2mm}
\end{figure}

\begin{table}[]
\caption{Results of model selection. $\texttt{SAT}$ indicates property violation
}
\vspace{-2mm}
\label{tab:formal_verification}
\begin{tabular}{cccccccc}
             &             & \multicolumn{4}{c}{\textbf{Safety Properties}}                                                            &                 &               \\
\multicolumn{2}{c|}{\textbf{}} &
  \multicolumn{1}{c|}{\textbf{$\Theta_\uparrow$}} &
  \multicolumn{1}{c|}{$\Theta_\downarrow$} &
  \multicolumn{1}{c|}{$\Theta_\leftarrow$} &
  \multicolumn{1}{c|}{$\Theta_\rightarrow$} &
  \multicolumn{2}{c}{\textbf{Model Selection}} \\ \hline
\multicolumn{2}{|c|}{\textbf{Method}} &
  \multicolumn{1}{c|}{$\texttt{SAT}$} &
  \multicolumn{1}{c|}{$\texttt{SAT}$} &
  \multicolumn{1}{c|}{$\texttt{SAT}$} &
  \multicolumn{1}{c|}{$\texttt{SAT}$} &
  \multicolumn{2}{c|}{\textit{Completely safe model}} \\ \hline
\multicolumn{2}{|c|}{PPO}  & \multicolumn{1}{c|}{300} & \multicolumn{1}{c|}{246} & \multicolumn{1}{c|}{80}  & \multicolumn{1}{c|}{167} & \multicolumn{2}{c|}{0}          \\
\multicolumn{2}{|c|}{L-PPO} & \multicolumn{1}{c|}{221} & \multicolumn{1}{c|}{198}  & \multicolumn{1}{c|}{53} & \multicolumn{1}{c|}{161} & \multicolumn{2}{c|}{\textbf{3}} \\ \hline
\end{tabular}
\vspace{-5mm}
\end{table}
According to the results reported in Table~\ref{tab:formal_verification}, L-PPO has fewer violations than PPO, confirming that incorporating soft constraints in training has a direct impact on decreasing violations of hard constraints. We show the positions of the hard constraints violation of PPO on one of the colon models in Fig.~\ref{fig:adversarial}c. As expected, large proportions of violations take place at sharp bendings, which are the critical points for the correct execution of the colonoscopic procedure.

This analysis sought to ascertain if it is feasible to identify a policy that adheres to all the hard constraints. As Table~\ref{tab:formal_verification} attests, three models satisfying all the hard constraints were identified in the case of L-PPO, while no policies conforming to the same standards were observed in the case of PPO, demonstrating the efficacy of the framework proposed herein. It is noteworthy that the L-PPO utilized in prior experiments (for example, in Table~\ref{tab:centerline_results}) is one of the three safe policies.


In order to emphasize the vulnerability of DNNs and the necessity of using FV in these safety critical scenarios, we can consider Fig.~\ref{fig:adversarial}. This figure displays the results of the analysis on a policy trained with PPO. Fig.~\ref{fig:adversarial}b shows an input on which the tested model acts safely, without violating the property, while with the same property $\Theta_\uparrow$ in Fig.~\ref{fig:adversarial}a, an adversarial input is discovered by the formal verifier. It is clear that the input only differs by 0.18 in the 6th value, yet this insignificant alteration causes the network to output a secure action in one case and a potentially dangerous action in the other.

\vspace{-2mm}
\section{CONCLUSIONS}  \label{sec:conclusions}

In this work, we investigate the challenges associated with the deployment of DRL for autonomous colonoscopy navigation in a virtual simulation. DRL-based methods have demonstrated the ability to successfully traverse patient-specific colon models with comparable performance to that of expert clinicians \cite{pore2022colonoscopy}. Nevertheless, these methods are susceptible to adversarial attacks, which could result in safety violations with potentially fatal consequences. Consequently, we exploit a CRL approach that ensures soft safety constraints through a cost function of safety violations. 
However, enforcing hard constraints through this methodology is difficult. 
To this end, we propose a model selection strategy that harnesses FV to evaluate the safety of a vast pool of policies trained using CRL. The FV is a modular framework capable of verifying any given set of safety properties and is able to provide guarantees of safe behavior prior to deployment. From the 300 policies trained using CRL, we identified three policies that adhered to all safety constraints, compared to no policies that met the same criterion for standard DRL.

In our future work, we aim to conduct real-robot experiments and extend the experimental validation under varying illumination conditions. We also plan to investigate the effects of abnormalities such as fluids, polyps and colon movements.








\bibliographystyle{IEEEtran}{

\bibliography{root}}

\begin{thebibliography}{10}
\providecommand{\url}[1]{#1}
\csname url@samestyle\endcsname
\providecommand{\newblock}{\relax}
\providecommand{\bibinfo}[2]{#2}
\providecommand{\BIBentrySTDinterwordspacing}{\spaceskip=0pt\relax}
\providecommand{\BIBentryALTinterwordstretchfactor}{4}
\providecommand{\BIBentryALTinterwordspacing}{\spaceskip=\fontdimen2\font plus
\BIBentryALTinterwordstretchfactor\fontdimen3\font minus
  \fontdimen4\font\relax}
\providecommand{\BIBforeignlanguage}[2]{{%
\expandafter\ifx\csname l@#1\endcsname\relax
\typeout{** WARNING: IEEEtran.bst: No hyphenation pattern has been}%
\typeout{** loaded for the language `#1'. Using the pattern for}%
\typeout{** the default language instead.}%
\else
\language=\csname l@#1\endcsname
\fi
#2}}
\providecommand{\BIBdecl}{\relax}
\BIBdecl

\bibitem{manfredi2021endorobots}
L.~Manfredi, ``Endorobots for colonoscopy: design challenges and available
  technologies,'' \emph{Front. Robot. AI}, vol.~8, p. 705454, 2021.

\bibitem{martin2020enabling}
J.~W. Martin \emph{et~al.}, ``Enabling the future of colonoscopy with
  intelligent and autonomous magnetic manipulation,'' \emph{Nature Mach.
  Intell.}, vol.~2, no.~10, pp. 595--606, 2020.

\bibitem{pore2023autonomous}
A.~Pore \emph{et~al.}, ``Autonomous navigation for robot-assisted intraluminal
  and endovascular procedures: A systematic review,'' \emph{IEEE Trans.
  Robot.}, 2023.

\bibitem{prendergast2018autonomous}
J.~M. Prendergast \emph{et~al.}, ``Autonomous localization, navigation and
  haustral fold detection for robotic endoscopy,'' in \emph{Proc. IEEE Int.
  Conf. Intell. Robot. Syst.}\hskip 1em plus 0.5em minus 0.4em\relax IEEE,
  2018, pp. 783--790.

\bibitem{prendergast2020real}
J.~M. Prendergast, G.~A. Formosa \emph{et~al.}, ``A real-time state dependent
  region estimator for autonomous endoscope navigation,'' \emph{IEEE Trans.
  Robot.}, vol.~37, no.~3, pp. 918--934, 2020.

\bibitem{reilink2010image}
R.~Reilink, S.~Stramigioli, and S.~Misra, ``Image-based flexible endoscope
  steering,'' in \emph{Proc. IEEE Int. Conf. Intell. Robot. Syst.}\hskip 1em
  plus 0.5em minus 0.4em\relax IEEE, 2010, pp. 2339--2344.

\bibitem{lazo2022autonomous}
J.~F. Lazo \emph{et~al.}, ``Autonomous intraluminal navigation of a soft robot
  using deep-learning-based visual servoing,'' in \emph{Proc. IEEE Int. Conf.
  Intell. Robot. Syst.}\hskip 1em plus 0.5em minus 0.4em\relax IEEE, 2022, pp.
  6952--6959.

\bibitem{pore2022colonoscopy}
A.~Pore \emph{et~al.}, ``Colonoscopy navigation using end-to-end deep
  visuomotor control: A user study,'' in \emph{Proc. IEEE Int. Conf. Intell.
  Robot. Syst.}\hskip 1em plus 0.5em minus 0.4em\relax IEEE, 2022, pp.
  9582--9588.

\bibitem{li2023rl}
K.~Li \emph{et~al.}, ``Rl-tee: Autonomous probe guidance for transesophageal
  echocardiography based on attention-augmented deep reinforcement learning,''
  \emph{IEEE Trans. Automat. Sci. Eng.}, 2023.

\bibitem{AcHeTa17}
J.~Achiam \emph{et~al.}, ``Constrained policy optimization,'' in \emph{Int.
  Conf. Mach. Learn.}, 2017.

\bibitem{MaCoFa21b}
E.~Marchesini, D.~Corsi, and A.~Farinelli, ``{Exploring Safer Behaviors for
  Deep Reinforcement Learning},'' in \emph{Proc. 35th AAAI Conf. on Artificial
  Intelligence (AAAI)}, 2021.

\bibitem{RaAcAm19}
A.~Ray, J.~Achiam, and D.~Amodei, ``Benchmarking safe exploration in deep
  reinforcement learning,'' \emph{arXiv preprint arXiv:1910.01708}, 2019.

\bibitem{LiuSurvey}
C.~Liu \emph{et~al.}, ``Algorithms for verifying deep neural networks,''
  \emph{Found. Trends{\textregistered} Optim.}, vol.~4, no. 3-4, pp. 244--404,
  2021.

\bibitem{ProVe}
D.~Corsi \emph{et~al.}, ``Formal verification of neural networks for
  safety-critical tasks in deep reinforcement learning,'' in \emph{Uncert.
  Artif. Intel.}\hskip 1em plus 0.5em minus 0.4em\relax PMLR, 2021, pp.
  333--343.

\bibitem{TACAS}
G.~Amir \emph{et~al.}, ``Verifying learning-based robotic navigation systems,''
  \emph{arXiv preprint arXiv:2205.13536}, 2022.

\bibitem{pore2021safe}
A.~Pore \emph{et~al.}, ``Safe reinforcement learning using formal verification
  for tissue retraction in autonomous robotic-assisted surgery,'' in
  \emph{Proc. IEEE Int. Conf. Intell. Robot. Syst.}\hskip 1em plus 0.5em minus
  0.4em\relax IEEE, 2021, pp. 4025--4031.

\bibitem{incetan2021vr}
K.~{\.I}ncetan \emph{et~al.}, ``Vr-caps: A virtual environment for capsule
  endoscopy,'' \emph{Med. Imag. Anal.}, vol.~70, p. 101990, 2021.

\bibitem{HaZhTu18}
T.~Haarnoja \emph{et~al.}, ``Soft actor-critic algorithms and applications,''
  \emph{arXiv preprint arXiv:1812.05905}, 2018.

\bibitem{schulman2017proximal}
J.~Schulman, F.~Wolski, P.~Dhariwal, A.~Radford, and O.~Klimov, ``Proximal
  policy optimization algorithms,'' \emph{arXiv preprint arXiv:1707.06347},
  2017.

\bibitem{CountingProVe}
L.~Marzari, D.~Corsi, F.~Cicalese, and A.~Farinelli, ``The \#dnn-verification
  problem: Counting unsafe inputs for deep neural networks,'' in \emph{IJCAI},
  2023.

\bibitem{Reluplex}
G.~Katz \emph{et~al.}, ``Reluplex: An efficient smt solver for verifying deep
  neural networks,'' in \emph{Int. conf. comp. verif.}\hskip 1em plus 0.5em
  minus 0.4em\relax Springer, 2017, pp. 97--117.

\bibitem{Verinet2}
P.~Henriksen \emph{et~al.}, ``Deepsplit: An efficient splitting method for
  neural network verification via indirect effect analysis.'' in \emph{IJCAI},
  2021, pp. 2549--2555.

\end{thebibliography}


\end{document}